\documentclass[10pt,twocolumn,letterpaper]{article}

\usepackage{iccv}
\usepackage{times}
\usepackage{graphicx}
\usepackage{amsmath}
\usepackage{amssymb}

\usepackage[T1]{fontenc}
\usepackage[utf8]{inputenc}
\usepackage{authblk}

\usepackage{float}
\usepackage{wrapfig}

\usepackage[labelfont=bf]{caption}
\usepackage{color}
\usepackage{mathtools} 
\usepackage{amsfonts} 

\usepackage{array}
\usepackage{booktabs}
\usepackage{rotating}
\usepackage{multirow}
\usepackage{multicol}
\usepackage{lscape}
\usepackage{subfig}

\usepackage[export]{adjustbox}

\usepackage{algorithm,algorithmicx,algpseudocode}

\usepackage{color}

\usepackage[pagebackref=true,breaklinks=true,letterpaper=true,colorlinks,bookmarks=false]{hyperref}

\usepackage[breaklinks=true,bookmarks=false]{hyperref}

\iccvfinalcopy 

\begin{document}

\title{Dynamic Multiscale Tree Learning Using Ensemble Strong Classifiers for Multi-label Segmentation of Medical Images with Lesions}

\author[1,2]{Samya Amiri}
\author[1]{Mohamed Ali Mahjoub}
\author[2]{Islem Rekik\thanks{Corresponding author: {irekik@dundee.ac.uk}, \url{www.basira-lab.com}} }
\affil[1]{LATIS lab, ENISo -- National Engineering School of Sousse, Tunisia }
\affil[2]{BASIRA lab, CVIP group, School of Science and Engineering, Computing, University of Dundee, UK }

\renewcommand\Authands{ and }

\maketitle

\begin{abstract}
   We introduce a dynamic multiscale tree (DMT) architecture that learns how to leverage the strengths of different state-of-the-art classifiers for supervised multi-label image segmentation. Unlike previous works that simply aggregate or cascade classifiers for addressing image segmentation and labeling tasks, we propose to embed strong classifiers into a tree structure that allows bi-directional flow of information between its classifier nodes to gradually improve their performances. Our DMT is a generic classification model that inherently embeds different cascades of classifiers while enhancing learning transfer between them to boost up their classification accuracies.  Specifically, each node in our DMT can nest a Structured Random Forest (SRF) classifier or a Bayesian Network (BN) classifier. The proposed SRF-BN DMT architecture has several appealing properties. First, while SRF operates at a patch-level (regular image region), BN operates at the super-pixel level (irregular image region), thereby enabling the DMT to integrate multi-level image knowledge in the learning process. Second, although BN is powerful in modeling dependencies between image elements (superpixels, edges) and their features, the learning of its structure and parameters is challenging. On the other hand, SRF may fail to accurately detect very irregular object boundaries. The proposed DMT robustly overcomes these limitations for both classifiers through the ascending and descending flow of contextual information between each parent node and its children nodes. Third, we train DMT using different scales for input patches and superpixels. Basically, as we go deeper along the tree edges nearing its leaf nodes, we progressively decrease the patch and superpixel sizes, producing segmentation maps that capture a coarse-to-fine image details. Last, DMT demonstrates its outperformance in comparison to several state-of-the-art segmentation methods for multi-labeling of brain images with gliomas.
\end{abstract}

\section{Introduction}

\begin{figure*}
\begin{center}
\includegraphics[scale=0.25]{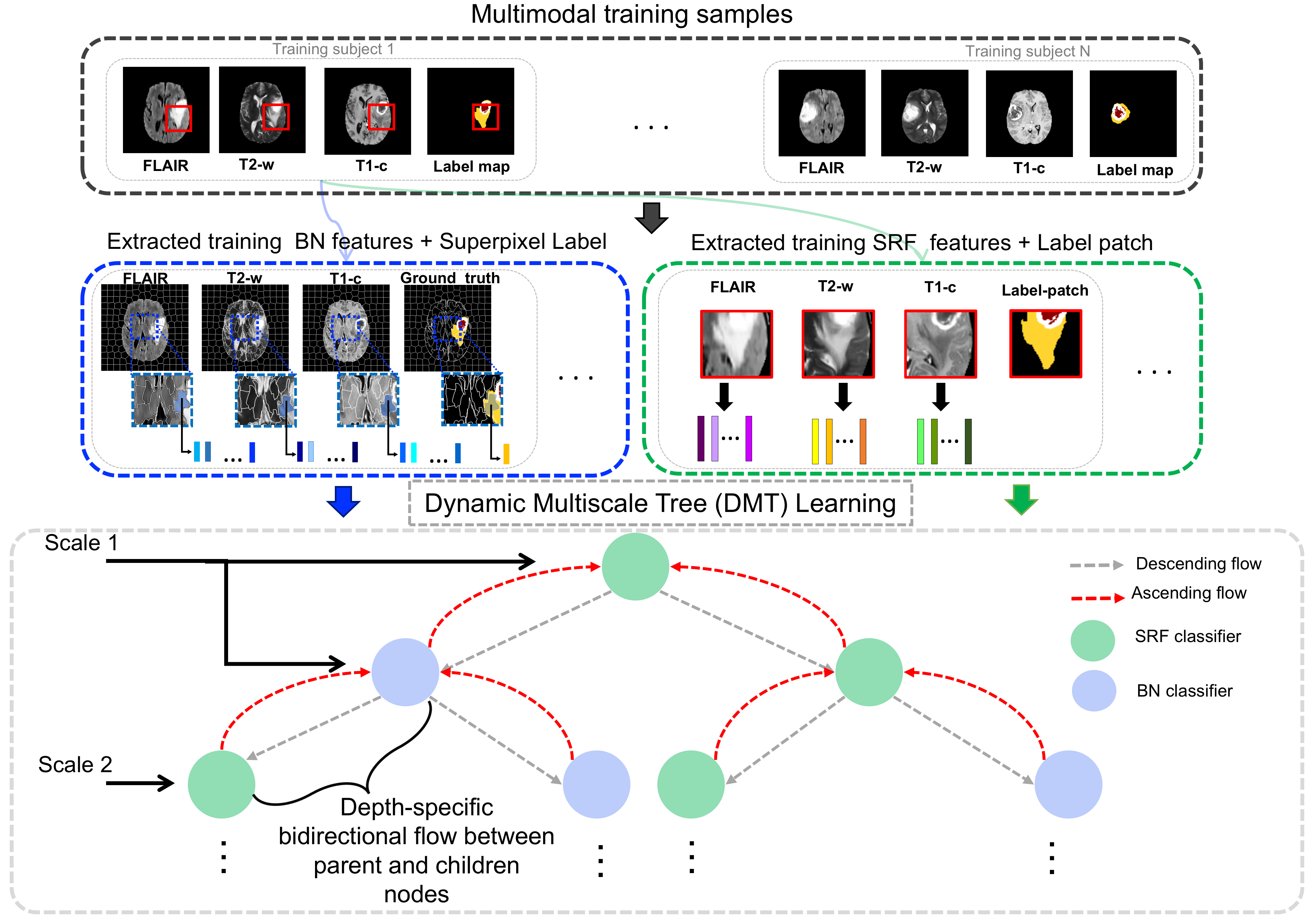}
\end{center}
   \caption{Proposed Dynamic Multi-scale Tree (DMT) learning architecture for multi-label classification (training stage). DMT embeds SRF and BN classifiers into a binary tree architecture, where a depth-specific bidirectional flow occurs between parent and children nodes making the tree learning dynamic. During training, SRF learns a mapping between feature patches extracted from three MRI modalities and their corresponding label patches for each training subject; whereas, BN classifier learns conditional probabilities from the superpixels of the oversegmented multimodal MR images and the label map.}
\label{Fig1}
\end{figure*}

Accurate multi-label image segmentation is one of the top challenges in both computer vision and medical image analysis. Specifically, in computer-aided healthcare applications, medical image segmentation constitutes a critical step for tracking the evolution of anatomical structures and lesions in the brain using neuroimaging, as well as quantitatively measuring group structural difference between image populations \cite{havaei2017brain,valverde2017improving,christ2017automatic,wang2016area,8}. Multi-label image segmentation is widely addressed as a classification problem. Previous works \cite{li2004multilabel,wei2014cnn} used individual classifiers such as support vector machine (SVM) to segment each label class independently, then fuse the different label maps into a multi-label map. However, prior to the fusion step, the produced label maps may largely overlap one another, which might yield to biased fused label map. Alternatively, the integration of multiple classifiers within the same segmentation framework would help reduce this bias and improve the overall multi-label classification performance since M heads are better than one as reported in \cite{lee2015m}. Broadly, one can categorize the segmentation methods that combine multiple classifiers into two groups:(1) cascaded classifiers, and(2) ensemble classifiers.
In the first group, classifiers are chained such that the output of each classifier is fed into the next classifier in the cascade to generate the final segmentation result at the end of the cascade. Such architecture can be adopted for two different goals. First, cascaded classifiers take into account contextual information, encoded in the segmentation map outputted from the previous classifier, thereby enforcing spatial consistency between neighboring image elements (e.g., patches, superpixels) in the spirit of an auto-context model \cite{havaei2017brain,tu2010auto,qian2016vivo,zhang2016concatenated, 8}. Second, this allows to combine classifiers hierarchically, where each classifier in the cascade is assigned to a more specific segmentation task (or a sub-task), as it further sub-labels the output label map of its antecedent classifier \cite{dai2016instance,valverde2017improving,christ2017automatic,wang2016area}. Although these methods produced promising results, and clearly outperformed the use of single (non-cascaded) classifiers in different image segmentation applications, cascading classifiers only allows a unidirectional learning transfer, where the learned mapping from the previous classifier is somehow ‘communicated' to the next classifier in the chain for instance through the output segmentation map.  
 The second group represents ensemble classifiers based methods, which train individual classifiers, then aggregate their segmentation results \cite{rahman2014ensemble}. Specifically, such frameworks combine a set of independently trained classifiers on the same labeling problem and generates the final segmentation result by fusing the individual segmentation results using a fusion method, which is typically weighted or unweighted voting \cite{kim2015randomized}. Hence, it constructs a strong classifier that outperforms each individual `weak' classifier (or base classifier) \cite{lee2015m}. For instance, Random Forest (RF) classification algorithm, independently trains weak decision trees using bootstrap samples generated from the training data to learn a mapping between the feature and the label sets \cite{breiman2001random}. The segmentation map of a new input image is the aggregation of the trees' decisions by major voting. RF demonstrated its efficiency in solving different image classification problems \cite{qian2016vivo,zhang2016concatenated}, which reflects the power of the ensemble classifiers technique. In addition to significantly improving the segmentation results when compared with single classifiers, ensemble classifiers based methods are powerful in addressing several known classification problems such as imbalanced correlation and over-fitting \cite{yijing2016adapted}. However, such combination technique is not enough to fully exploit the training of classifiers and leverage their strengths. Indeed, the base classifiers perform segmentation independently without any cooperation to solve the target classification problem. Moreover, the learning of each classifier in the ensemble is performed in one-step, as opposed to multi-step classifier training, where the learning of each classifier gradually improves from one step to the next one. We note that this differs from cascaded classifiers, where each classifier is ‘visited' or trained once through combining the contextual segmentation map of the previous classifier along with the original input image.

To address the aforementioned limitations of both categories, we propose a Dynamic Multi-scale Tree (DMT) architecture for multi-label image segmentation. DMT is a binary tree, where each node nests a classifier, and each traversed path from the root node to a leaf node encodes a cascade of classifiers (i.e., nodes on the path). Our proposed DMT architecture allows a bidirectional information flow between two successive nodes in the tree (from parent node to child node and from child node back to parent node). Thus, DMT is based on ascending and descending feedbacks between each parent node and its children nodes. This allows to gradually refine the learning of each node classifier, while benefitting from the learning of its immediate neighboring nodes. To generate the final segmentation results, we combine the elementary segmentation results produced at the leaf nodes using majority voting strategy.  The proposed architecture integrates different possible combinations of different classifiers, while taking advantage of their strengths and overcoming their limitations through the bidirectional learning transfer between them, which defines the dynamic aspect of the proposed architecture. Furthermore, the DMT inherently integrates contextual information in the classification task, since each classifier inputs the segmentation result of its parent node or children nodes classifiers. Additionally, to capture a coarse-to-fine image details for accurate segmentation, the DMT is designed to consider different scale at each level in the tree in a way that the adopted scale decreases as we go deeper along the tree edges nearing its leaf nodes. 

In this work, we define our DMT classification model using two strong classifiers: Structured Random Forest (SRF) and Bayesian Network (BN). SRF is an improved version of Random Forest \cite{kontschieder2011structured}. In addition of being fast, resistant to over-fitting and having a good performance in classifying high-dimensional data, SRF handles structural informationand integrates spatial information. It has shown good performance in several classification tasks especially muli-label image segmentation \cite{kontschieder2011structured,zhang2016segmentation}. On the other hand, BN is a learning graphical model that statistically represents the dependencies between the image elements and their features. It is suitable for multi-label segmentation for its effectiveness in fusing complex relationships between image features of different natures and handling noisy as well as missing signals in images \cite{zhang2008integration,panagiotakis2011natural,zhang2011bayesian,yang2015multi}. Embedding SRF and BN within our DMT leverages their strengths and helps overcome their limitations (i.e. not accurately classifying transitions between label classes for SRF and the problem of parameters learning such as prior probabilities for BN). Moreover, the SRF-BN bidirectional cooperation during learning and testing stages enables the integration of multi-level image knowledge through the combination of regular and irregular image elements (i.e. patch-level classification produced by SRF and superpixel-level classification produced by BN).
To sum up, our SRF-BN DMT has promise for multi-label image segmentation as it:
\begin{itemize}
	\item Gradually improves the classification accuracy through the bidirectional flow between parents and children nodes, each nesting a BN or SRF classifier
\end{itemize}
\begin{itemize}
	\item Simultaneously integrates multi-level and multi-scale knowledge from training images, thereby examining in depth the different inherent image characteristics
\end{itemize}
\begin{itemize}
	\item Overcomes SRF and BN limitations when used independently through multiple cascades (or tree paths) composed of different combinations of BN and SRF classifiers.
\end{itemize}

\section{Base classifiers}
In this section we briefly introduce the SRF and BN classifiers, that are embedded as nodes in our DMT classification framework. Then,we explain in detail how we define our DMT architecture and elaborate on how to perform the training and testing stages on an image dataset for multi-label image segmentation.

\subsection{Structured Random Forest}

SRF is a variant of the traditional Random Forest classifier, which better handles and preserves the structure of different labels in the image \cite{kontschieder2011structured}. While, standard RF maps an intensity feature vector extracted from a 2D patch centered at pixel $\emph{x}$ to the label of its center pixel $\emph{x}$  (i.e., patch-to-pixel mapping), SRF maps the intensity feature vector to a 2D label patch centered at $\emph{x}$  (patch-to-patch mapping). This is achieved at each node in the SRF tree, where the function that splits patch features between right and left children nodes depends on the joint distribution of two labels: a first label at the patch center   and a second label selected at a random position within the training patch \cite{kontschieder2011structured}. We also note that in SRF, both feature space and label space nest patches that might have different dimensions. Despite its elegant and solid mathematical foundation as well as its improved performance in image segmentation compared with RF, SRF might perform poorly at irregular boundaries between different label classes since it is trained using regularly structured patches \cite{kontschieder2011structured}. Besides, it does not include contextual information to enforce spatial consistency between neighboring label patches. To address these limitations, we first propose to embed SRF as a classifier node into our DMT architecture, where the contextual information is provided as a segmentation map by its parent and children nodes. Second, we improve its training around irregular boundaries through leveraging the strength of one or more its neighboring BN classifiers, which learns to segment the image at the superpixel level, thereby better capturing irregular boundaries in the image.

\subsection{Bayesian network}

Various BN-based models have been proposed for image segmentation \cite{zhang2008integration,panagiotakis2011natural,zhang2011bayesian}. In our work, we adopt the BN architecture proposed in \cite{zhang2011bayesian}. As a preprocessing step, we first generate the edge maps from the input MR image modalities Fig\ref{Fig1}. This edge map consists of a set of superpixels ${S_{p_{i}}} ; i=1 \dots N$ (or regional blobs) and edge segments ${E_{j}};  j=1 \dots L$. 

We define our BN as a four-layer network, where each node in the first layer stores a superpixel. The second layer is composed of nodes, each storing a single edge from the edge map. The two remaining layers store the extracted superpixel features and edge features, respectively. During the training stage, to set BN parameters, we define the prior probability $P(S_{p_{i}})$ of $S_{p_{i}}$  as a uniform distribution and then learn the conditional probability  representing the relationship between the superpixels' features and their corresponding labels using a mixture of Gaussians model. In addition, we empirically define the conditional probability modeling the relationship between each superpixel label and each edge state (i.e., true or false edge)  $P(E_{j}|p_a (E_{j}))$, where  $p_a (E_{j})$  denotes the parent superpixel nodes of $E_{j}$. 

During the testing stage, we learn the BN structure through encoding the semantic relationships between superpixels and edge segments. Specifically, each edge node has for parent nodes the two superpixel nodes that are separated by this edge. In other words, each superpixel provides contextual information to judge whether the edge is on the object boundary or not. If two superpixels have different labels, it is more likely that there is a true object boundary between them, i.e. $E_{j} = 1$, otherwise $E_{j} = 0$ .

Although automatic segmentation methods based on BN have shown great results in the state-of-the-art, they might perform poorly in segmenting low-contrast image regions and different regions with similar features  \cite{zhang2011bayesian}. To further improve the segmentation accuracy of BN, we propose to include additional information through embedding BN classifier into our proposed DMT learning architecture.

\section{Proposed Multi-scale Dynamic Tree Learning}
In this section, we present the main steps in devising our Multi-scale Dynamic Tree segmentation framework, which aims to boost up the performance of classifiers nested in its nodes. Fig\ref{Fig1}  illustrates the proposed binary tree architecture composed of classifier nodes, where each classifier ultimately communicates the output of its learning (i.e., semantic context or probability segmentation maps) to its parent and children nodes. Therefore, the learning of the tree is dynamic as it is based on ascending and descending feedbacks between each parent node and its children nodes. Specifically, each node output is fed to the children nodes as semantic context, in turn the children nodes transfer their learning (i.e. probability maps) to their common parent node. Then, after merging these transferred probability maps from children nodes, the parent node uses the merged maps as a contextual information to generate a new segmentation result that will be subsequently communicated again to its children nodes. This gradually improves the learning of its classifier nodes at each depth level of the tree. In the following sections, we further detail the DMT architecture.

\subsection{Dynamic Tree Learning}
\begin{figure}[t]
\begin{center}
\includegraphics[width=0.8\linewidth]{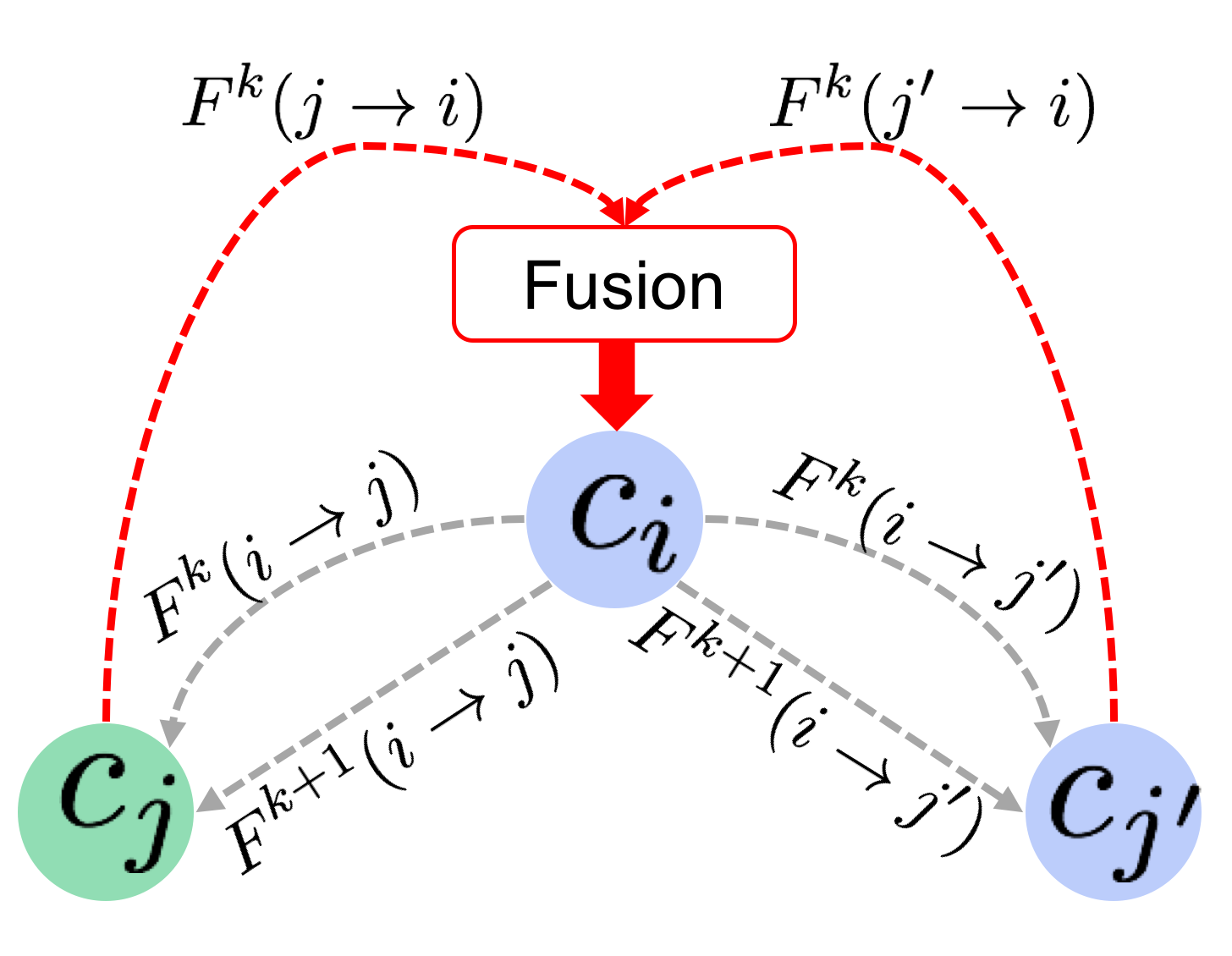}
\end{center}
   \caption{\emph{Implicit} and \emph{explicit}learning transfer in the proposed dynamic multi-scale tree-based classification architecture.. The dashed gray arrows denote the descending flows from the parent classifier $c_i$ to its children  $c_j$ and $c_j'$ (iteration $k$), while the dashed red arrows denote the ascending flows derived from the children nodes to their parent node. Both ascending flows are fused for the parent node to integrate them in generating a new segmentation map that will be communicated to its two children classifier nodes (iteration $k+1$).}
\label{Fig2}
\label{fig:onecol}
\end{figure}

We define a binary tree T(V,E), where V denotes the set of nodes in T and E represents the set of edges in T. Each node $i$ in T represents a classifier $c_i$ and each edge $e_{ij}$ connecting two nodes i and j carries bidirectional contextual information flow between the classifiers $c_i$ and $c_j$ that are always inputting the original image characteristics (i.e. the features for SRF, superpixel features and input image edge map for BN). Specifically, we define bidirectional feedbacks between two neighboring   classifier nodes $i$ and $j$, encoding two flows: a descending flow $F_{i\rightarrow j} $ that represents the transfer of the probability maps generated by parent classifier node $c_i$ to its child classifier node $c_j$ as contextual information and an ascending flow $F_{j\rightarrow i} $ that models the transfer of the probability maps generated by a child node $c_j$ back to its parent node $c_i$ (Fig.~\ref{Fig2}). This depth-wise bidirectional learning transfer occurs locally along each edge between a parent node and its child node, thereby defining the dynamics of the tree. 

In addition, as our Dynamic Tree (DT) grows exponentially, it integrates various combinations of classifiers. Thus, each path of the tree implements a unique  cascade of classifiers. To generate the final segmentation result we aggregate the segmentation maps produced at each leaf node in the binary tree by applying the majority voting. 

\textbf{Inherent implicit and explicit transfer learning between nodes in DT architecture.} We note that the bidirectional flow between parent nodes and their corresponding children nodes defines a new traversing strategy of the tree nodes, that in addition to the dynamic learning aspect, encodes two different types of learning transfer: explicit and implicit. Indeed, the ascending and descending flows between parent nodes and their children through the direct transfer of their generated probability maps is an explicit learning transfer. However, in our binary tree, when a parent node $i$ receives the ascending flows ($F^k(j \rightarrow i)$ and $F^k(j' \rightarrow i)$ from its left and right children nodes j and j', they are fused before being passed on, in a second round, as contextual information ($F^{k+1}(i \rightarrow j)$ and $F^{k+1}(i \rightarrow j') $) to the children nodes (Fig.~\ref{Fig2}). The probability maps fusion at the parent node level is performed through simple averaging. In particular, the parent node concatenates the fused probability map with the original input features to generate a new segmentation probability map result that will be communicated to its two children classifier nodes. Hence, the children nodes of the same parent node explicitly cooperate to improve their parent learning, and implicitly cooperate to improve their own learning while using their parent node as a proxy.

\textbf{SRF-BN Dynamic Tree.} In this work, each classifier node is assigned a SRF or a BN model, previously described in Section 2, to define our Dynamic Tree architecture. The transferred information between classifiers through the descending and ascending flows is used in addition to the testing image features as contextual  information, while BN classifier uses this information as prior knowledge (i.e prior probability) to perform the multi-label segmentation task. The combination of SRF and BN classifiers is compelling for the following reasons. First, it enhances the performance of BN by taking the posterior probability generated by SRF as prior probability. This justifies our choice of the root node of our DT as a SRF.  Second, it improves SRF performance around irregular between-class boundaries since SRF benefits from BN structure learning, which is based on image over-segmentation that is guided by object boundaries. Third, as the SRF maps image information at the patch level, while BN models knowledge at the superpixel level, their combination allows the aggregation of regular (i.e. patch) and irregular (i.e. superpixel) structures in the image for our target multi-label segmentation task.

\subsection{Dynamic Multi-scale Tree Learning}
To further boost the performance of our multi-label segmentation framework and enhance the segmentation accuracy, we introduce a multi-scale learning strategy in our dynamic tree architecture by varying the size of the input patches and superpixels used to grow the SRF and construct the BN classifier. Specifically, we use a different scale at each depth level such as we go deeper along the tree edges nearing its leaf nodes, we progressively decrease the size of both patches and superpixels in the training and testing stages.
In addition to capturing coarse-to-fine details of the image anatomical structure, the application of the multi-scale strategy to the  proposed DT allows to capture fine-to-coarse information. Indeed, DMT learning semantically divides the image into different patterns (e.g., different patches and superpixels at each depth of the tree) in both intensity and label domains at different scales. However, thanks to the bidirectional dynamic flow, the scale defined at each depth influences the performance of parent nodes (in previous level) and children nodes (in next level), which allows to simultaneously perform coarse-to-fine and fine-to-coarse information integration in the multi-label classification task. Moreover, a depth-wise multi-scale feature representation adaptively encodes image features at different scales for each image pixel in the image element (superpixel or patch).

\subsection{Statistical superpixel-based and patch-based feature extraction}
To train each classifier node in the tree, we extract the following statistical features at the superpixel level (for BN) and 2D patch level (for SRF): first order operators (mean, standard deviation, max, min, median, Sobel, gradient); higher order operators (Laplacian, difference of Gaussian, entropy, curvatures, kurtosis, skewness), texture features (Gabor filter), and spatial context features (symmetry, projection, neighborhoods) \cite{prastawa2004brain}.


\section{ Results and Discussion}

\textbf{Dataset and parameters.} We evaluate our proposed brain tumor segmentation framework on 50 subjects with high-grade gliomas, randomly selected from the Brain Tumor Image Segmentation Challenge (BRATS 2015) dataset \cite{menze2015multimodal}. For each patient, we use three MRI modalities (FLAIR, T2-w, T1-c) along with the corresponding manually segmented glioma lesions. They are rigidly co-registered and resampled to a common resolution to establish patch-to-patch correspondence across modalities. Then, we apply N4 filter for inhomogeneity correction, and use histogram linear transformation for intensity normalization.
\begin{table*}
\begin{center}
\begin{tabular}{c|c|c|c}
\hline
Methods & HT&	CT&	ET \\
\hline\hline
Dynamic Multiscale Tree-Learning (depth =2)&	\textbf{\textcolor{blue}{89.64}} &	\textbf{\textcolor{blue}{82.3}}	& \textbf{\textcolor{blue}{80}} \\
Dynamic Multiscale Tree-Learning (depth =1)	& \textbf{88.93} &	\textbf{79.7}	& \textbf{78.09}\\
\hline
Dynamic Tree-Learning (depth =2)&	\textbf{89.56}	& \textbf{80.43}	& \textbf{79}\\
Dynamic Tree-Learning (depth =1)	& \textbf{88.07} &	\textbf{78.7} &	\textbf{77.94}\\
	\hline		
SRF-BN&	82.5	&72.6&	70\\
SRF-SRF&	80	&70.05	&37.12\\
BN-BN&	79.82	&71	&56. 14\\
\hline
SRF&	75&	60&	35\\
BN	&70.8	&45&	32\\
\hline
\end{tabular}
\end{center}
\caption{Segmentation results of the proposed framework and comparison methods averaged across 50 patients. (HT: whole Tumor; CT: Core Tumor; ET: Enhanced Tumor; depth of the tree; * indicates outperformed methods with $p-value \prec0.05$).}
\label{table}
\end{table*}

For the baseline methods training we adopt the following parameters:(1) Edgemap generation: we use the SLICE oversegmentation algorithm with a superpixel number fixed to 1000 and compactness fixed to 10 \cite{achanta2010slic}. To establish superpixel-to-superpixel correspondence across modalities for each subject, we first  oversegment the FLAIR MRI, then we apply the generated edgemap (i.e., superpixel partition) to the corresponding T1-c and T2-w MR images. (2) SRF training: we grow 15 trees using intensity feature patches of size 10x10 and label patches of size 7x7. (3) BN construction: the BN model is built using the generated edgemap as detailed in Section 2; the conditional probabilities modeling the relationships between the superpixel labeling and the edge state are defined as follows: $P(E_{j}=1|p_a (E_{j}))= 0.8$ if the parent region nodes have different labels, and $P(E_{j}=1|p_a (E_{j})) = 0.2$ otherwise.

\textbf{Evaluation and comparison methods.} For comparison, as baseline methods we use: (1) SRF: the Random Forest version that exploits structural information described in Section 2, (2) BN: the classification algorithm described in Section 2 where the prior probability of superpixels is set as a uniform distribution, (3) SRF-SRF denotes the auto-context Structured Random Forest, (4) BN-BN denotes the auto-context Bayesian Network, where the first BN prior probability is set as a uniform distribution while the second classifier use the posterior probability of its previous as prior probability. Of note, by conventional auto-context classifier, we mean a uni-directional contextual flow from one classifier to the next one.
The segmentation frameworks were trained using leave-one-patient cross-validation experiments. For evaluation, we use the Dice score between the ground truth region area $A_{gt}$ and the segmented region area $A_s$ as follows $D = (A_{gt} \bigcap A_s)/ 2( A_{gt} + A_s)$. 

Next, we investigate the influence of the tree depth as well as the multi-scale tree learning strategy on the performance of the proposed architecture.

\textbf{Varying tree architectures. }In this experiment, we evaluate two different tree architectures to examine the impact of the tree depth on the framework performance. Table.~\ref{table} shows the segmentation results for 2-level tree (i.e. depth=2) and 1-level tree (i.e. depth=1) for tumor lesion multi-label segmentation with and without multiscale variant. Although the average Dice Score has improved from depth 1 to 2, the improvement wasn't statistically significant. We did not explore larger depths (d>2), since as the binary tree grows exponentially, its computational time dramatically increases and becomes demanding in terms of resources (especially memory). 

\textbf{Multi-scale tree architecture.} To examine the influence of the multiscale DT learning strategy, we compare the conventional DT architecture (at a fixed-scale) to MDT architecture. For the fixed-scale architecture, all tree nodes nest either an SRF classifier trained using intensity feature patches of size 10x10 and label patches of size 7x7  or a BN classifier constructed using an edgemap of 1000 superpixels generated with a compactness of 10. In the multiscale architecture, we keep the same parameters of the fixed-scale architecture at the first level of the tree while the classifiers of the second level are trained with different parameters. Specifically, we use smaller intensity patches (of size 8x8) and label patches (of size 5x5) for the SRF training, and a smaller number of superpixels for BN construction (1200 superpixels).

\begin{figure*}
\begin{center}
\includegraphics[scale=0.25]{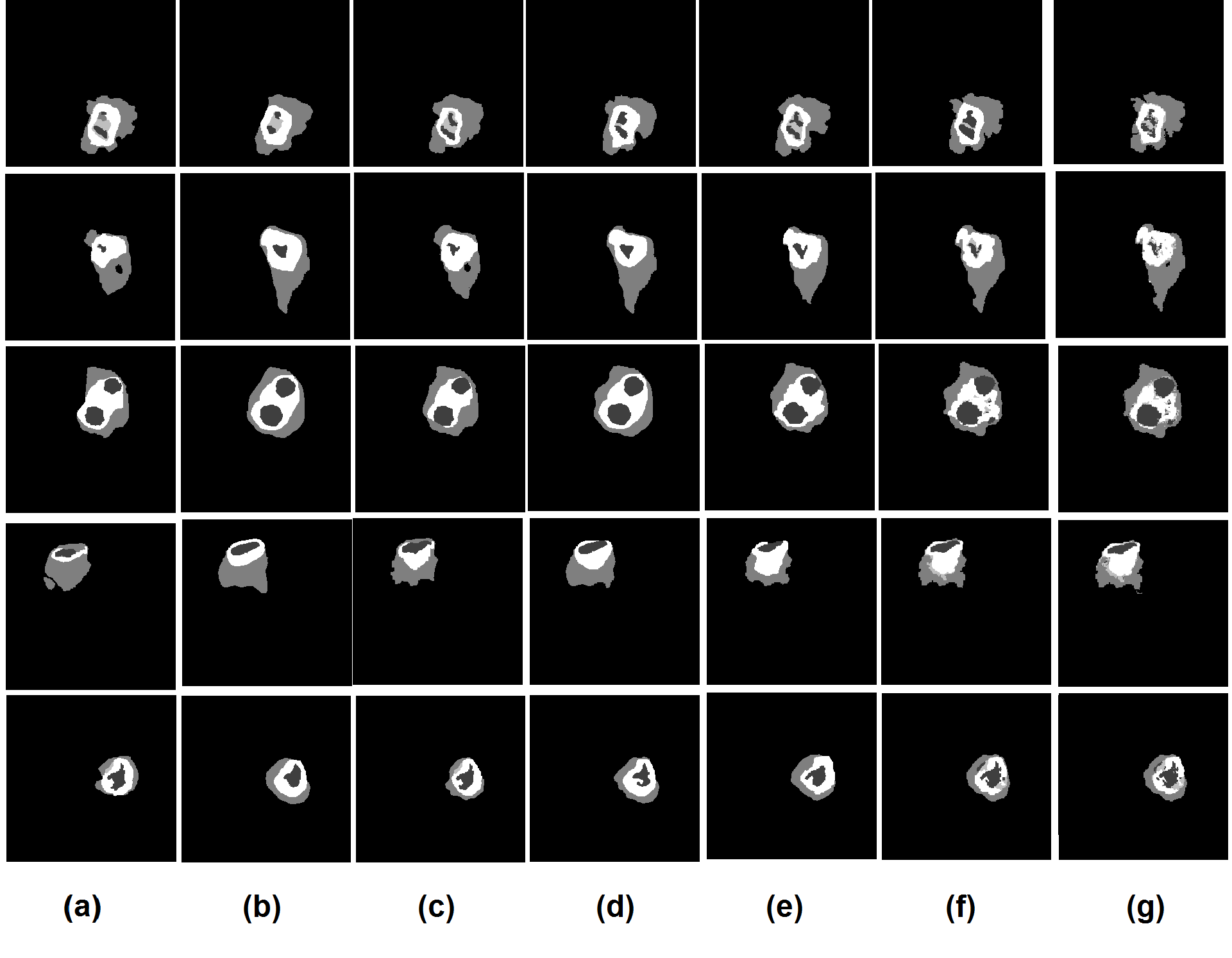}
\end{center}
   \caption{Qualitative segmentation results for all the baseline methods applied on 5 subjects: (a) BN segmentation result. (b) SRF segmentation result. (c) Auto-context BN. (d) Auto-context SRF. (e) SRF+BN segmentation result. (f) DMT segmentation result. (depth=2). (g) Ground truth label map.}
\label{Fig3}
\end{figure*}


Clearly, the quantitative results show the outperformance (improvement of 7\%) of both proposed DT  and DMT architectures in comparison with several baseline methods for multi-label tumor lesion segmentation with statistical significance (p < 0.05) . This indicates that a deeper combination of different learning models helps increase the segmentation accuracy.
When comparing the results of the SRF and BN we found that SRF outperforms BN in segmenting the three classes: wHole Tumor (HT), Core Tumor (CT) and Enhancing Tumor (ET) Table.~\ref{table} . This is due to the fact that BN have difficulties in segmenting low-contrast images and identifying different superpixels having similar characteristics, especially with the lack of any prior knowledge on the anatomical structure of the testing image.	Although BN has a low Dice score compared to SRF, in Fig.~\ref{Fig3} we can note that it has better performance in detecting the boundaries between different classes. This shows the impact of the irregular structure of superpixels used during BN training and testing, which gives BN the ability to be more accurate in detecting object boundaries compared to SRF that considers regular image patches. Notably, BN structure is individualized during the testing stage for each testing subject since it is based on the testing image oversegmentation map. Thus, SRF and BN classifiers are complementary. First, they perform segmentation at regular and irregular structures of the image. Second, one (SRF) learns image knowledge during the training stage, while the other (BN) is structured using the input testing image during the testing stage through modeling the testing  image structure.
Further, the results of SRF-SRF and BN-BN models that implement the auto-context approach show an improvement of the segmentation results at both qualitative and quantitative levels when compared with baseline SRF and BN models. More importantly, we note that BN-BN cascade outperforms SRF-SRF cascade when segmenting the Core Tumor and Enhancing Tumor (ET) lesions. This can be explained by the fine and irregular anatomical details of these image structures when compared to the whole tumor lesion. Since BN is trained using irregular superpixels, it produced more accurate segmentations for these classes (e.g., BN-BN:56.14 vs SRF-SRF: 37.12 for ET).
Through further cascading both SRF and BN classifiers, we note that the heterogenoues SRF-BN cascade produced much better results compared to both autocontext SRF and autocontext BN for two main reasons. First SRF aids in defining BN prior based on the testing image structure, while BN enhances the performance of SRF at the boundaries level. This further highlights the importance of integrating both regular and irregular image elements for training classifiers that capture different image structures. 
The outperformance of the proposed DMT architecture also lays ground for our assumption that embedding SRF and BN into our a unified dynamic architecture where they mutually benefit from their learning boosts up the multi-label segmentation accuracy. In addition to the previously mentionned advantages of combining SRF and BN, it is important to note that the integration of variant cascades of SRF and BN endows our architecture with a an efficient learning ability, where it incorporates in a deep manner the knowledge of SRF based on modeling the dataset during the training step and the individualized learning of BN based on the testing image during the testing step.
Notably, our DMT architecture has a few limitations. First, it becomes more demanding in terms of computational and memory resources, as the tree grows exponentially (in the order of $O(2^n)$). The more nodes we add to the binary tree, the slower the algorithm converges. Second, the patch and superpixel sizes in the multi-scale learning strategy can be further learned, instead of empirically fixing them through inner cross-validation. Third, the bidirectional flow is currently restricted between neighboring parent and children nodes at a fixed tree depth. This can be extended to further nodes (e.g., root node), where the semantic context progressively diffuses from each node $i$ along tree paths to far-away nodes.


\section{Conclusion}

We proposed a Dynamic Multi-scale Tree (DMT) learning architecture that both cascades and aggregates classifiers for multi-label medical image segmentation. Specifically, our DMT embeds classifiers into a binary tree architecture, where each node nests a classifier and each edge encodes a learning transfer between the classifiers. A new tree traversal strategy is proposed where a depth-wise bidirectional feedbacks are performed along each edge between a parent node and its child node. This allows explicit learning between parent and children nodes and implicit learning transfer between children of the same parent. Moreover, we train DMT using different scales for input patches and superpixels to capture a coarse-to-fine image details as well as a fine-to-coarse image structures through the depth-wise bidirectional flow. To sum up, our DMT integrates compound and complementary  aspects: deep learning, cooperative learning, dynamic learning, coarse-to-fine and fine-to-coarse learning. In our future work, we will devise a more comprehensive tree traversal strategy where the learning transfer starts from the root node, descending all the way down to the leaf nodes and then ascending all the way up to the root node. We will also evaluate our DMT semantic segmentation architecture on different large datasets.

{\small
\bibliographystyle{ieee}
\bibliography{egbib}

\begin{thebibliography}{10}\itemsep=-1pt

\bibitem{achanta2010slic}
R.~Achanta, A.~Shaji, K.~Smith, A.~Lucchi, P.~Fua, and S.~S{\"u}sstrunk.
\newblock Slic superpixels.
\newblock Technical report, 2010.

\bibitem{breiman2001random}
L.~Breiman.
\newblock Random forests.
\newblock {\em Machine learning}, 45(1):5--32, 2001.

\bibitem{christ2017automatic}
P.~F. Christ, F.~Ettlinger, F.~Gr{\"u}n, M.~E.~A. Elshaera, J.~Lipkova,
  S.~Schlecht, F.~Ahmaddy, S.~Tatavarty, M.~Bickel, P.~Bilic, et~al.
\newblock Automatic liver and tumor segmentation of ct and mri volumes using
  cascaded fully convolutional neural networks.
\newblock {\em arXiv preprint arXiv:1702.05970}, 2017.

\bibitem{dai2016instance}
J.~Dai, K.~He, and J.~Sun.
\newblock Instance-aware semantic segmentation via multi-task network cascades.
\newblock In {\em Proceedings of the IEEE Conference on Computer Vision and
  Pattern Recognition}, pages 3150--3158, 2016.

\bibitem{havaei2017brain}
M.~Havaei, A.~Davy, D.~Warde-Farley, A.~Biard, A.~Courville, Y.~Bengio, C.~Pal,
  P.-M. Jodoin, and H.~Larochelle.
\newblock Brain tumor segmentation with deep neural networks.
\newblock {\em Medical image analysis}, 35:18--31, 2017.

\bibitem{kim2015randomized}
H.~Kim, J.~Thiagarajan, J.~Jayaraman, and P.-T. Bremer.
\newblock A randomized ensemble approach to industrial ct segmentation.
\newblock In {\em Proceedings of the IEEE International Conference on Computer
  Vision}, pages 1707--1715, 2015.

\bibitem{kontschieder2011structured}
P.~Kontschieder, S.~R. Bulo, H.~Bischof, and M.~Pelillo.
\newblock Structured class-labels in random forests for semantic image
  labelling.
\newblock In {\em Computer Vision (ICCV), 2011 IEEE International Conference
  on}, pages 2190--2197. IEEE, 2011.

\bibitem{lee2015m}
S.~Lee, S.~Purushwalkam, M.~Cogswell, D.~Crandall, and D.~Batra.
\newblock Why m heads are better than one: Training a diverse ensemble of deep
  networks.
\newblock {\em arXiv preprint arXiv:1511.06314}, 2015.

\bibitem{li2004multilabel}
X.~Li, L.~Wang, and E.~Sung.
\newblock Multilabel svm active learning for image classification.
\newblock In {\em Image Processing, 2004. ICIP'04. 2004 International
  Conference on}, volume~4, pages 2207--2210. IEEE, 2004.

\bibitem{8}
L.~F. Loic, V.~N. Aditya, A.-V. Javier, L.~Richard, and C.~Antonio.
\newblock Segmentation of brain tumors via cascades of lifted decision forests.
\newblock In {\em Proceedings of BRATS Challenge-MICCAI}, 2016.

\bibitem{menze2015multimodal}
B.~H. Menze, A.~Jakab, S.~Bauer, J.~Kalpathy-Cramer, K.~Farahani, J.~Kirby,
  Y.~Burren, N.~Porz, J.~Slotboom, R.~Wiest, et~al.
\newblock The multimodal brain tumor image segmentation benchmark (brats).
\newblock {\em IEEE transactions on medical imaging}, 34(10):1993--2024, 2015.

\bibitem{panagiotakis2011natural}
C.~Panagiotakis, I.~Grinias, and G.~Tziritas.
\newblock Natural image segmentation based on tree equipartition, bayesian
  flooding and region merging.
\newblock {\em IEEE Transactions on Image Processing}, 20(8):2276--2287, 2011.

\bibitem{prastawa2004brain}
M.~Prastawa, E.~Bullitt, S.~Ho, and G.~Gerig.
\newblock A brain tumor segmentation framework based on outlier detection.
\newblock {\em Medical image analysis}, 8(3):275--283, 2004.

\bibitem{qian2016vivo}
C.~Qian, L.~Wang, Y.~Gao, A.~Yousuf, X.~Yang, A.~Oto, and D.~Shen.
\newblock In vivo mri based prostate cancer localization with random forests
  and auto-context model.
\newblock {\em Computerized Medical Imaging and Graphics}, 52:44--57, 2016.

\bibitem{rahman2014ensemble}
A.~Rahman and S.~Tasnim.
\newblock Ensemble classifiers and their applications: A review.
\newblock {\em arXiv preprint arXiv:1404.4088}, 2014.

\bibitem{tu2010auto}
Z.~Tu and X.~Bai.
\newblock Auto-context and its application to high-level vision tasks and 3d
  brain image segmentation.
\newblock {\em IEEE Transactions on Pattern Analysis and Machine Intelligence},
  32(10):1744--1757, 2010.

\bibitem{valverde2017improving}
S.~Valverde, M.~Cabezas, E.~Roura, S.~Gonz{\'a}lez-Vill{\`a}, D.~Pareto, J.-C.
  Vilanova, L.~Rami{\'o}-Torrent{\`a}, {\`A}.~Rovira, A.~Oliver, and
  X.~Llad{\'o}.
\newblock Improving automated multiple sclerosis lesion segmentation with a
  cascaded 3d convolutional neural network approach.
\newblock {\em arXiv preprint arXiv:1702.04869}, 2017.

\bibitem{wang2016area}
L.~Wang, P.~Pedersen, E.~Agu, D.~Strong, and B.~Tulu.
\newblock Area determination of diabetic foot ulcer images using a cascaded
  two-stage svm based classification.
\newblock {\em IEEE Transactions on Biomedical Engineering}, 2016.

\bibitem{wei2014cnn}
Y.~Wei, W.~Xia, J.~Huang, B.~Ni, J.~Dong, Y.~Zhao, and S.~Yan.
\newblock Cnn: Single-label to multi-label.
\newblock {\em arXiv preprint arXiv:1406.5726}, 2014.

\bibitem{yang2015multi}
S.~Yang, C.~Yuan, B.~Wu, W.~Hu, and F.~Wang.
\newblock Multi-feature max-margin hierarchical bayesian model for action
  recognition.
\newblock In {\em Proceedings of the IEEE Conference on Computer Vision and
  Pattern Recognition}, pages 1610--1618, 2015.

\bibitem{yijing2016adapted}
L.~Yijing, G.~Haixiang, L.~Xiao, L.~Yanan, and L.~Jinling.
\newblock Adapted ensemble classification algorithm based on multiple
  classifier system and feature selection for classifying multi-class
  imbalanced data.
\newblock {\em Knowledge-Based Systems}, 94:88--104, 2016.

\bibitem{zhang2016segmentation}
J.~Zhang, Y.~Gao, S.~H. Park, X.~Zong, W.~Lin, and D.~Shen.
\newblock Segmentation of perivascular spaces using vascular features and
  structured random forest from 7t mr image.
\newblock In {\em International Workshop on Machine Learning in Medical
  Imaging}, pages 61--68. Springer, 2016.

\bibitem{zhang2008integration}
L.~Zhang and Q.~Ji.
\newblock Integration of multiple contextual information for image segmentation
  using a bayesian network.
\newblock In {\em Computer Vision and Pattern Recognition Workshops, 2008.
  CVPRW'08. IEEE Computer Society Conference on}, pages 1--6. IEEE, 2008.

\bibitem{zhang2011bayesian}
L.~Zhang and Q.~Ji.
\newblock A bayesian network model for automatic and interactive image
  segmentation.
\newblock {\em IEEE Transactions on Image Processing}, 20(9):2582--2593, 2011.

\bibitem{zhang2016concatenated}
L.~Zhang, Q.~Wang, Y.~Gao, H.~Li, G.~Wu, and D.~Shen.
\newblock Concatenated spatially-localized random forests for hippocampus
  labeling in adult and infant mr brain images.
\newblock {\em Neurocomputing}, 2016.

\end{thebibliography}
}

\end{document}